\documentclass[letterpaper, 10 pt, conference]{ieeeconf} \IEEEoverridecommandlockouts                            

\IEEEoverridecommandlockouts                            
\usepackage{graphicx}
\usepackage{float}
\usepackage{atbegshi}
\usepackage{afterpage}
\usepackage{stfloats}
\usepackage{amsmath}
\usepackage{subfigure}
\usepackage{amssymb}
\usepackage{tabularx}
\usepackage{comment}

\usepackage{algorithm}
\usepackage{algpseudocode}
\usepackage{hyperref}

\usepackage[dvipsnames]{xcolor}
\usepackage{caption}

\usepackage{colortbl}

\newcommand\lp[1]{{\textcolor{Green}{#1}}}

\usepackage[style=ieee]{biblatex}

\title{\LARGE \bf
Distributed Multi-robot Source Seeking in Unknown Environments with Unknown Number of Sources
}

\author{Lingpeng Chen$^{1}$, Siva Kailas$^{3}$, Srujan Deolasee$^{2}$, Wenhao Luo$^{4}$, Katia Sycara$^{2}$ and Woojun Kim$^{2}$
\thanks{$^{1}$Lingpeng Chen is with Shenzhen Institute of Artificial Intelligence and Robotics for Society at the Chinese University of Hong Kong, Shenzhen. ${\tt\small \{lingpengchen\}@link.cuhk.edu.cn}$}
\thanks{$^{2}$Woojun Kim, Srujan Deolasee, and Katia Sycara are with the Robotics Institute at Carnegie Mellon University. ${\tt\small\{woojunk,sdeolase,sycara\}@andrew.cmu.edu}$}
\thanks{$^{3}$Siva Kailas is with the School of Interactive Computing at Georgia Institute of Technology ${\tt\small \{skailas3\}@gatech.edu}$}
\thanks{$^{4}$Wenhao Luo is with the Department of Computer Science at University of Illinois Chicago. ${\tt\small \{wenhao\}@uic.edu}$}
}  
\begin{document}

\maketitle

\begin{abstract}
    We introduce a novel distributed source seeking framework, \texttt{DIAS}, designed for multi-robot systems in scenarios where the number of sources is unknown and potentially exceeds the number of robots. Traditional robotic source seeking methods typically focused on directing each robot to a specific strong source and may fall short in comprehensively identifying all potential sources. \texttt{DIAS} addresses this gap by introducing a hybrid controller that identifies the presence of sources and then alternates between exploration for data gathering and exploitation for guiding robots to identified sources. It further enhances search efficiency by dividing the environment into Voronoi cells and approximating source density functions based on Gaussian process regression. Additionally, \texttt{DIAS} can be integrated with existing source seeking algorithms. We compare \texttt{DIAS} with existing algorithms, including \texttt{DoSS} and \texttt{GMES} in simulated gas leakage scenarios where the number of sources outnumbers or is equal to the number of robots. 
    The numerical results show that \texttt{DIAS} outperforms the baseline methods in both the efficiency of source identification by the robots and the accuracy of the estimated environmental density function.
    
\end{abstract}

\section{Introduction}
    \label{Sec:Intro}
    Large-scale information gathering is considered crucial in real-world scenarios, such as search and rescue \cite{lee2018receding, liu2013robotic}, environmental monitoring \cite{chen2019pareto, Kailas2023, luo2018adaSamp, gadipudi2024offripp}, and wildlife monitoring \cite{tokekar2013tracking, deolasee2024dypnipp}. Given the demanding nature of these tasks, there is a pressing need for robotic automation to aid in information collection. One notable area within this domain is source seeking, where robots are deployed in an unknown environment to find sources of interest. Its applications include tasks such as chemical spills investigation \cite{chemical_investigation} \cite{DoSS}, signal emitter identification \cite{TRO_AdaSearch}, and light source seeking \cite{BO, TRO_light_source}. 
    The typical approach for this involves defining a quantity function that indicates the degree of source interest and then optimizing the target positions of multiple robots to maximize this quantity function \cite{DoSS}. 
    However, from a practical perspective, previous studies have a notable limitation: they are unable to reliably locate all potential sources, particularly when the number of sources outnumbers the robots. This is because existing approaches mainly concentrate on directing each robot to the strongest source \cite{BO, fast_and_scale}, or to a limited subset of the most significant sources \cite{DoSS, TRO_AdaSearch}.
    In many practical scenarios, robots need to continuously search for sources, as the number of sources is generally unknown. 
    
\begin{figure}[t]
        \centering
        \includegraphics[width=9cm]{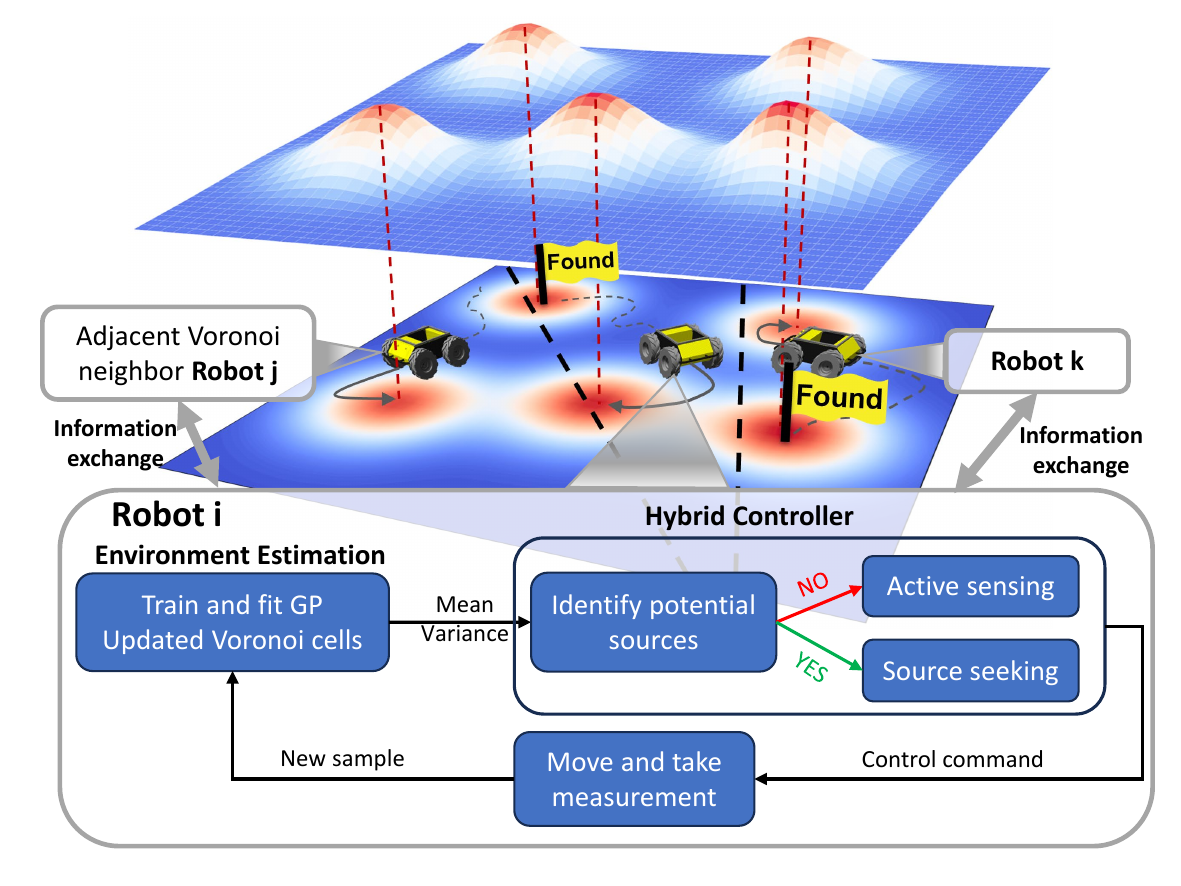}
        \caption{Overview: three robots navigate in a field to identify all potential sources. The upper layer shows the distribution of the gas concentration level. The lower layer depicts the unexplored 2D environment with a projected heat map of the distribution. The black dashed lines represent Voronoi partition boundaries. }
        \label{fig:flowchart}
        \vspace{-3ex}
    \end{figure}

    
    In this paper, we particularly focus on the multi-robot source seeking problem, where there are multiple static sources in an unknown environment and the number of sources is unknown, requiring multiple robots to continuously search for sources. This problem is illustrated in Fig. \ref{fig:flowchart}. Here, the robot team is deployed into an unknown area and is expected to cooperatively find multiple static local-maxima sources, where the sources outnumber the robots. The main difference from the problem addressed in prior works is that the robot needs to keep finding other sources once it has located one. To the best of our knowledge, there is no existing literature that effectively addresses the challenge of employing the multi-robot system to find all potential sources, particularly in scenarios where the number of sources outnumber the number of robots.

    To address this problem, we propose a new \textbf{D}istributed source seeking framework that enables multiple robots to cooperatively \textbf{I}dentify \textbf{A}ll potential \textbf{S}ources, named as \texttt{DIAS}. To enable multiple robots to continuously search for sources, we introduce a hybrid controller that identifies potential sources and then conducts either active sensing — aimed at gathering additional informative data (exploration) — or source seeking — intended to guide robots to the identified potential source (exploitation), based on the presence of some potential source. 
    Besides the hybrid controller, \texttt{DIAS} involves environment estimation that (1) divides the environment into Voronoi cells for efficient source seeking, and (2) approximates the unknown density function of sources from collected samples via Gaussian process regression. The proposed environment estimation encourages multiple robots to efficiently and cooperatively search space for sources. Another advantage of the proposed framework is its integrability, allowing it to be integrated with existing source seeking algorithms. We provide an example of how the proposed framework can be integrated with existing work, along with the corresponding result. 
    We compare the proposed framework, \texttt{DIAS}, with existing multi-robot source seeking algorithms, including \texttt{DoSS} \cite{DoSS}, \texttt{GMES} \cite{ma2023gaussian}, in two simulated gas leakage environments: one where the number of sources exceeds the number of robots, and another where it is equal. 
    The numerical results show that \texttt{DIAS} outperforms the considered baselines in terms of both the efficiency of source identification and the accuracy of the estimated environmental density function. In addition, we show that integrating \texttt{DIAS} with existing work yields improved results.

\section{Preliminaries} 
\subsection{Problem Setup}
    \label{Sec:Problem_setup}
    In this section, we formalize a new problem of multi-robot multi-source seeking, where the goal is to find all potential sources. Each source represents a point of emission, such as a gas leak  or chemical release, that generates a continuous scalar field in the environment. Let us consider a bounded two-dimensional (2D) environment $\mathcal{Q} \subseteq \mathbb{R}^{2}$ with $M$ static sources. Here, a set of positions is denoted by $\mathcal{S}$, and $s_k \in \mathcal{S}$ is the $k$-th source position. There exists an unknown \textit{density function} $\phi(\cdot):\mathcal{Q} \rightarrow \mathbb{R}_+$ that maps each position $q$ into a positive scalar value $\phi(q)$ indicating the degree of source emission. 
    We assume that $\phi$ is continuous over $\mathcal{Q}$ and has compact support within $\mathcal{Q}$. The robots are aware of the boundaries of $\mathcal{Q}$ but have no prior knowledge of the sources' locations and the specific form of the density function $\phi$.
    
    In this environment, a set of $N$ robots aims to identify all sources in the shortest amount of time. We assume that each robot measures the environmental phenomena by taking samples through some concentration sensors at a certain frequency with the following stochastic measurement model:
        \begin{equation}
            \hat{\phi} (x_i)=\phi(x_i)+\epsilon_i  
            \label{eq:measurement_model}
        \end{equation}
    where $x_i \in \mathcal{Q}$ and $\epsilon_i$ are the robot $i$'s location and the measurement noise. 
    Based on the measurements, each robot, employing only its local data and the limited information received from its neighbors, independently computes its goal for each iteration, aiming to identify the potential sources.
   
    
    Note that our objective is to identify all sources, whereas prior works focus on enabling multiple robots to locate sources, which limits their ability to identify all sources when the number of sources is unknown. Robots thus need to continue searching instead of staying when a source is found since the number of sources can exceed the number of robots (\(M>N\)). For this, we need to propose the specific criterion for source identification as follows:\\
    \textbf{Source identification criterion:} A source \(s_j\) is considered \textit{found} by a robot \(i\) if the robot's sampling location \(x_i\) is within a distance \(d\) from the source, satisfying the condition \(\left \| x_i - s_j \right \| \le d\). This definition is grounded in practical field explorations where robots employ various sensors in addition to the concentration sensor.


    \subsection{Gaussian Process Regression} \label{GP}
    
        Gaussian Process (GP) Regression is a non-parametric Bayesian approach to infer continuous functions by assuming that the target functions follow a multivariate joint Gaussian distribution \cite{seeger2004gaussian}. In this paper, we use GP regression to approximate the unknown density function. 
        That is, we find the mean function $\mu(q)=\mathbb{E}[\phi(q)]$  and the covariance (kernel) function $k(q,q') = \mathbb{E}[(\phi(q)-\mu(q))^T(\phi(q')-\mu(q'))]$ and then model each position $q$ as a Gaussian distribution of $\phi(q)\sim \mathcal{N}(\mu(q), \sigma^2(q))$.
        Here, we use squared-exponential kernel function $k(q, q') = \sigma_f^2 \exp\left(-(q - q')^T (q - q')/(2l^2)\right)$, where the hyper-parameters $l$ and $\sigma_f$ are length-scale and scale factor \cite{rasmussen2006gaussian}. We optimize the hyper-parameters of the GP model based on the training data \( \mathbf{X} =  [x_1, \cdots, x_{t}]^T \) and the corresponding measurements \( \mathbf{y} =  [\hat{\phi} (x_1), \cdots, \hat{\phi} (x_t)]^T \). We then use these parameters to infer the posterior mean $\mu_{q_{test}|\mathbf{X},\mathbf{y}}$ and variance $\sigma^2_{q_{test}|\mathbf{X},\mathbf{y}}$ of the test data point $q_{test}$ as follows:
        \begin{align} \label{eq:gp}
                \mu(q_{\text{test}}) &= \mathbf{k}(q_{\text{test}})^T (\mathbf{K}_{\mathbf{X}} + \sigma_n^2\mathbf{I})^{-1}\mathbf{y}, \\
                \sigma^2(q_{\text{test}}) &= k(q_{\text{test}}, q_{\text{test}}) - \mathbf{k}(q_{\text{test}})^T (\mathbf{K}_{\mathbf{X}} + \sigma_n^2\mathbf{I})^{-1} \mathbf{k}(q_{\text{test}}), \nonumber
        \end{align}
        where \(\mathbf{k}(q_{\text{test}}) = [k(x_1, q_{\text{test}}), \cdots, k(x_t, q_{\text{test}})]^T\) and $\mathbf{K}_{\mathbf{X}}$ is the positive definite symmetric kernel matrix $[k(q, q')]_{q, q' \in \mathbf{X} \cup q_{\text{test}}}$. Here, the hyperparameters that we optimize are $\theta = (\sigma_n, \sigma_f, l)$. These are trained by maximizing the log of the marginal likelihood function as follows.
        \begin{equation}
            \begin{aligned}
                \theta^* &= \arg \max_{\theta} \log p(\mathbf{y} |\mathbf{X}, \theta) \\
                &= -\frac{1}{2}\mathbf{y}^T \mathbf{K_X}^{-1} \mathbf{y} - \frac{1}{2}\log|\mathbf{K_X}| - \frac{n}{2}\log 2\pi.
            \end{aligned}
            \label{eq:gp_train}
        \end{equation}
        By training the hyper-parameters to their optimal values, we adjust the structure of the covariance function to best represent the sampled data. 

\subsection{Ergodic Control} \label{sec:ergodic}
Ergodic control is a coverage control algorithm that relies on the concept of trajectory ergodicity \cite{miller2015ergodic, abraham2018decentralized}. A trajectory is ergodic with respect to a distribution if its time-averaged statistics match the distribution’s spatial statistics. In information gathering task, it aims to provide an informative path so that the amount of time the robot spends in a region is proportional to the expected information in that region. An ergodic trajectory is obtained by optimizing the control command $u$ to iteratively minimize ergodic metric $\mathcal{E}$, which measures how far a robot trajectory is from being ergodic with respect to the expected information density (EID). In a $2D$ space, we first consider the time-averaged statistics $c(x(t))$ for the robot's trajectory $x(t)$\footnote{Here, for simplicity and readability, we leave out the $i$th indexing notation for robot $i$.} during the time interval $t \in [0, T]$, written as $c(x(t))=\frac{1}{T} \int_{0}^{T} \delta\left(x-x(\tau)\right) d\tau$, where $\delta$ is the Dirac function. Then, the ergodic metric is defined as, 
\begin{equation}
    \label{eq:ergodic_metric}
    \mathcal{E}(x(t))=\sum_{k \in \mathbb{N}^2} \lambda_k\left(\mathcal{I}_k - c_k(x(t)) \right)^2, 
\end{equation}
\noindent 
where $\mathcal{I}_k$ and $c_k$ are the $k$-th Fourier coefficients of the EID function $I(\cdot)$ and trajectory statistics $c(x(t))$. 
 $\lambda_k=\left(1+\|k\|^2\right)^{-\frac{3}{2}}$ in the ergodic metric denotes the weight for each Fourier coefficient. Here, the EID function that we considered will be described in Section \ref{subsec:hybrid}. By optimizing this ergodic metric, a closed-form solution for low-level robot control can be obtained~\cite{abraham2018decentralized}. Here, the robots are assumed to be operated based on a nonlinear control affine system. Due to space limitations, we refer to the solution provided in prior work~\cite{abraham2018decentralized}.

\section{Proposed Algorithm}

    In this section, we introduce a novel framework that enables multiple robots to make strategic decisions between exploration and exploitation by assessing the presence of additional potential sources, thereby identifying all possible sources. 
    Our framework involves three main steps: (1) updating Voronoi cells based on robot locations and estimating the environment based on self-collected samples, (2) identifying the potential sources and calculating the control command using the hybrid controller (either to active sensing or source seeking), and (3)  moving and taking a new measurement from the environment. When a robot identifies a source using a predefined criterion, it declares it and informs neighboring robots, ensuring all are aware of discovered sources to avoid repeated visits. 
    The algorithm is outlined in Alg. 1, with each component elaborated in the subsequent sections.

    \subsection{Environment Estimation}
        \label{SubSec:environment_estimation}
     At each iteration $t$, we first divide the environment into $N$ Voronoi cells, denoted as $V^t = \{V_1^t, \dots, V_N^t\}$, with one cell assigned to each robot $i$.
    The Voronoi tessellation is determined by the location of each robot $x_i$ and the positions of its nearby neighbors $x_j$ (where $j \ne i$). Each $V_i^t$ represents the region closest to robot $i$, defined as:
    \begin{equation}
        V_i^t = \{q \in Q: \|q - x_i\| \le \|q - x_j\|, \forall j \ne i\}
        \label{eq:voronoi}
    \end{equation}
    where $\|\cdot\|$ denotes the $l_2$-norm. The rationale behind this approach is to allocate each robot a responsible region to conduct environment estimation and source seeking task, which can ensure comprehensive area coverage and minimize overlap of robots' trajectories, as considered in the literature on multi-robot information gathering \cite{Multi_robot_coordination, luo2018adaSamp, luo2019distributed}. Note that each robot begins without prior knowledge of the environment, relying solely on initial observations obtained through sampling, as defined by Eq. \ref{eq:measurement_model}.
    

    We now estimate the unknown density function $\phi(q)$ over the considered map by training a GP model. Let us consider a robot $i$. At each iteration $t$, it obtains a measurement $\hat{\phi} (x_i^t)$ from its new sampling location $x_i^t$, which will be added to the local dataset $\mathbf{y}_i, \mathbf{X}_i$ respectively. 
    Based on the dataset, each robot $i$ trains the GP model using Eq. \ref{eq:gp_train} and use it for the hybrid controller. 

    \begin{algorithm} [t]
        \label{Algorithm:DIAS}
        \begin{algorithmic}[1] 
        \caption{DIAS}
        \For{each iteration $t = 1, 2, \dots$}
            \State $V^t = \{V_1^t, \dots, V_N^t\}$
            \Statex\hspace{3em} $\leftarrow \texttt{update\_voronoi}(x_1^t, \dots, x_N^t)$ \{Eq: \ref{eq:voronoi}\} 
            \For{each robot $i$ \textbf{simultaneously}}
                \State $y_i^t = \hat{\phi} (x_i^t) \leftarrow \texttt{take\_sample}(x_i^t)$  
                
                \State $\mathbf{X}_i, \mathbf{y}_i \leftarrow \mathbf{X}_i \cup x_i^t, \mathbf{y}_i \cup y_i^t$
                
                \State $\mu_i(\cdot), \sigma_i^2(\cdot) \leftarrow \texttt{GP}
                (\mathbf{X}_i, \mathbf{y}_i)$  \{Sec. \ref{SubSec:environment_estimation}\}
                
                \State $I_i(\cdot) \leftarrow \texttt{EID}(\mu_i(\cdot), \sigma_i^2(\cdot))$ 
                \{Eq. \ref{Eq:EID}\}

                \State $\mathcal{I}_{k,i} \leftarrow I_i(\cdot) $ \{Fourier inverse transform\}
                
                \State $\bar{\mathcal{I}}_k, \bar{c}_k \leftarrow \texttt{neighbour\_consensus}(\mathcal{I}_{k,i}, c_{k,i}, V^t)$ 
                \Statex\hspace{2.5em} \{Eq. \ref{eq:ergodic_consensus}\}
               
                \State $ \textit{is\_src}, q_{target} \leftarrow$ 
                \Statex\hspace{3em}$\texttt{identify\_potential\_source}(\mu_i(\cdot), \sigma_i^2(\cdot))$
                \Statex\hspace{2.5em} \{Sec. \ref{SubSec:potential_src_identification}\}

                \If{\textit{is\_src}}
                    \State $u_i \leftarrow \texttt{source\_seeking}(q_{target})$ 
                    \Statex\hspace{4em} \{Sec. \ref{SubSec:source_seeking_module}\}
                \Else
                    \State $u_i \leftarrow \texttt{active\_sensing}(\bar{\mathcal{I}}_k, \bar{c}_k)$ 
                    \Statex\hspace{4em} \{Sec. \ref{SubSec:active_sensing_module}\}
                \EndIf
                
                \State $x_i^{t+1} \leftarrow \texttt{robot\_dynamic\_model}(u_i)$ 
                \State $\texttt{source\_identification}(x_i^{t+1}, d)$
                \State \parbox[t]{\dimexpr\linewidth-\algorithmicindent}{Share discovered source location with neighbors\strut}
            \EndFor
            
        \EndFor
        \end{algorithmic}
    \end{algorithm}

    \subsection{Hybrid Controller} \label{subsec:hybrid}
        We propose a hybrid controller that identifies potential sources and enables the robot to navigate toward the potential source (i.e., exploitation) or move to gather information (i.e., exploration). 
        The proposed hybrid controller consists of three modules: (1) a potential source identification module, (2) an active sensing module, which aims to gather informative samples to improve the estimates of environmental distribution, and (3) a source seeking module, which aims to verify and navigate toward the source position.


    \subsubsection{Potential source identification module} \label{SubSec:potential_src_identification}
        In this module, each robot, within its designated Voronoi cell $V_i^t$, determines the presence of potential sources by analyzing the estimated distribution characterized by $\mu_i(\cdot)$ and $\sigma_i^2(\cdot)$. Given that sources represent extremum points in the distribution $\phi$, a robot can estimate the source's location by pinpointing local maxima in the estimated distribution $\mu_i(\cdot)$. Consequently, each robot can identify all potential source locations, denoted as $q_{pot}$, within its Voronoi cell $V_i^t$. For this, 
        we utilize the Lower Confidence Bound (LCB) criterion, defined as
        \begin{equation}\label{eq:lcb}
            LCB(q_{pot}) = \mu_i(q_{pot}) -\beta\sigma_i^2(q_{pot}),
        \end{equation}
        where $\beta$ is the width of the confidence bound. We propose to ascertain the presence of a potential source, denoted by \textit{is\_src}, as follows.
        \begin{align}
             \textit{is\_src} = \left\{
            \begin{array}{ll}
            True & \text{if } \exists ~ LCB(q_{pot}) > \tau ~\mbox{where}~ q_{pot} \in V_i^t \\
            False & \text{otherwise.}
            \end{array}\nonumber
            \right.
        \end{align} Here, $\tau$ is a threshold. Note that the rationale of using LCB is grounded in its effectiveness at prioritizing points that not only exhibit high estimated intensity—hinting at the possible presence of a source—but also feature low uncertainty, thereby increasing the reliability of the estimate. This dual capacity of the LCB enables more dependable differentiation between actual sources and noise. If \textit{is\_src} is true, the robot enters the source seeking module and selects the most confident $q_{pot}$ with the highest LCB value as the target $q_{target}$; otherwise, it proceeds to the active sensing.
        

    \subsubsection{Source seeking module} \label{SubSec:source_seeking_module}

        In this module, the robot navigates to the target source identified in the source identification module. To this end, we select target position $q_{target}$ and compute a low-level robot control command using Model Predictive Control (MPC), which is well-suited to manage the robot's dynamics and the constraints on control input \cite{maciejowski2002predictive}. Due to page constraints, the intricacies of the low-level control computation are not included.

        

    \subsubsection{Active sensing module}  \label{SubSec:active_sensing_module}
        If no potential sources are identified, the robot enters active sensing mode to gather more data and deepen its understanding of the environment. For this, we adopt ergodic control, which operates on the principle that an informative path should spend time in a region proportional to the amount of expected information in that region \cite{ErgodicOptimality}. Ergodic control necessitates a criterion for defining informativeness. Thus, we introduce an Expected Information Density (EID) function based on mutual information as this criterion, followed by a description of how ergodic control operates according to it. 

        Mutual Information quantifies the reduction of the overall uncertainty of the estimation when a new sample is acquired, but it is costly to calculate \cite{contal2014gaussian}. To avoid this, we calculate its lower bound $\gamma$ instead. Inspired by \cite{contal2014gaussian}, we use the lower bound of mutual information between the collected samples and the density function $\phi(\cdot)$ to calculate our EID function $I(q)$, which is computed as:
        \begin{equation}
            \label{Eq:EID}
            I(q) = \mu(q) + \alpha (\sqrt{\sigma^2(q) + \gamma_{t-1}} - \sqrt{\gamma_{t-1}}),
        \end{equation}
        where $\mu(\cdot)$ and $\sigma(\cdot)$ are obtained from the learned GP model, $\alpha$ is the exploration coefficient and $\gamma$ is the lower bound that is iteratively updated by $\gamma_t = \gamma_{t-1} + \sigma^2(x_i^t)$, in which $x_i^t$ is the newly sampled location at iteration $t$. 
        
        We now modify the extension of ergodic control to a multi-robot setting, as proposed in a prior work \cite{abraham2018decentralized}. From Sec. \ref{sec:ergodic}, each robot $i$ requires the Fourier coefficient $c_{k, i}$ and $\mathcal{I}_{k,i}$
        to compute the ergodic metric. Rather than using the individual metrics of each robot, we implement a multi-robot ergodic control that calculates control command based on the trajectory and the EID function of the overall multi-agent system, i.e., a common value of $\bar{c}_k$ and $\bar{\mathcal{I}}_k$, which are computed as 
       
        \begin{align}\label{eq:avgergodic}
            \bar{c}_k &= \frac{1}{N} \sum_{i=1}^{N} c_{k,i}, ~~~~
            \bar{\mathcal{I}}_k= \frac{1}{N} \sum_{i=1}^{N} \mathcal{I}_{k,i},
        \end{align}
        where $\mathcal{I}_{k,i}$ and $c_{k,i}$ represent the Fourier coefficients of the EID function and the trajectory information for each individual robot $i$. 
        This computation requires the exchange of local variable $\mathcal{I}_{k,i}$ and $c_{k,i}$ between the robots.
        In settings where robots can only communicate limited information to neighbors, we achieve Eq. \ref{eq:avgergodic} through a consensus-based method. The consensus is depicted as follows:
        \begin{align}\label{eq:ergodic_consensus}
            \lim_{t_c \to \infty} \sum_j P_{ij}^{t_c} c_{k,j} = \frac{1}{N} \sum_j c_{k,j} = \bar{c}_k,
        \end{align}
        with consensus matrix $P$ reflects the network's connectivity among agents, which is determined by the Voronoi tessellation $V^t$. \lp{} Robots exchange information $\mathcal{I}_{k,j}$ and $c_{k,j}$ with neighboring agents. The term $\sum_j P_{ij} c_{k,j}$ means robot $i$ receives $c_{k,j}$ from its Voronoi neighbor $j$ and take the average. The variable $t_c$ denotes the number of times the process has been conducted. With sufficient repetitions, agents reach a consensus, converging to the collective average statistics \( \bar{c}_k\). The consensus of $\bar{\mathcal{I}}_k$ is done in the same way. In our experiment, we iterated until $t_c = 5$ to achieve efficient computation.

        Now, each agent can get their optimal ergodic control command from the ergodic metric in Eq. \ref{eq:ergodic_metric}. Given that the dynamical state of each robot is independent, as noted in \cite{abraham2018decentralized}, the control policy $u_i$ for each robot can be deduced separately. This decentralized process is advantageous because the frequency information used is both compact and invariant over time, which reduces the communication load.

\section{Experimental Results}
    \label{Sec:Experiment}

    This section presents the simulation results in environments where gas dispersion is dominated by diffusion, forming Gaussian-shaped distributions. Assuming the diffusion speed is negligible relative to the robot's search time, the environment is considered static. This results in symmetric concentration gradients, typical of indoor or sheltered settings. The experiment includes two scenarios: (1) where the number of sources matches the number of robots, and (2) where the number of sources exceeds the number of robots.
   
    We compare our \texttt{DIAS} to three baselines: distributed online source seeking (\texttt{DoSS}) algorithm  \cite{DoSS}, Gaussian max-value entropy search (\texttt{GMES}) algorithm \cite{ma2023gaussian}, and a greedy information maximization algorithm developed based on Bayesian Optimization (BO) (\texttt{GreedyBO})~\cite{ma2023gaussian} framework, where agents move toward the highest UCB value within their Voronoi cell, with the UCB coefficient empirically set to 3 to ensure efficient performance across all scenarios. Here, \texttt{GMES} adopts a centralized approach within a multi-agent Bayesian optimization framework, while \texttt{DoSS} provides a distributed method and uses a Kalman consensus filter to model the distribution. All methods were tested under the same conditions, except \texttt{GMES}, which uses a centralized approach. It is important to note that \texttt{DIAS} is a general framework that can be applied to existing source seeking algorithms. To show this, we include \texttt{DIAS\_with\_GMES}, which replaces the source seeking module with \texttt{GMES}.

    \subsection{Simulated Environment Setup}
        We consider a simple $10m \times 10m$ map where three robots are deployed to find several sources. Here, we consider three scenarios: one with $3$ static sources (\textit{Scenario 1}), one with $5$ sources (\textit{Scenario 2}) and  and one with $7$ static sources (\textit{Scenario 3}). We conducted $30$ experiments for each algorithm in both scenarios respectively, varying the initial positions of the robots and sources in each trial. 
        Additionally, the density function $\phi(\cdot)$ is a superposition of Gaussian distributions, with each source's location corresponding to a peak, as derived from diffusion equation \cite{morton2019revival}. The intensity of each source $\phi(s)$ ranges from $0.16$ to $0.20$.

    \subsection{Performance Comparison}
        \label{SubSec:PerformanceAnalysis}

        
        \begin{table}[t]
        \caption{Comparison of Iterations Required to Identify All Sources (30 trials); lower values indicate better performance.}
        \label{tab:iterations}
        \centering
        \renewcommand{\arraystretch}{1.3}  
        \footnotesize     
        \setlength{\tabcolsep}{4pt}  
        \begin{tabular}{l|c|c|c}
        \hline
        \multicolumn{4}{c}{Number of iterations (mean $\pm$ standard deviation)} \\
        \hline
        Algorithm & 3 sources & 5 sources & 7 sources \\
        \hline
        DIAS (Ours) & \textbf{50.3$\pm$9.8} & 62.5$\pm$11.2 &  \textbf{69.4$\pm$14.6} \\
        DIAS\_with\_GMES (Ours) & 56.8$\pm$14.7 & \textbf{57.6$\pm$13.2} & 70.7$\pm$11.8 \\
        GreedyBO & 55.7$\pm$14.7 & 78.6$\pm$13.3 & 103.2$\pm$13.8
        \\
        DoSS & 110.0$\pm$25.1 & 163.4$\pm$36.1 & 194.7$\pm$28.4 \\
        GMES & 171.6$\pm$77.0 & 201.5$\pm$61.1 & 268.4$\pm$89.8 \\
        \hline
        \end{tabular}
        \end{table}

        \begin{figure*}[tp]
            \centering
            \includegraphics[width=17.5cm]{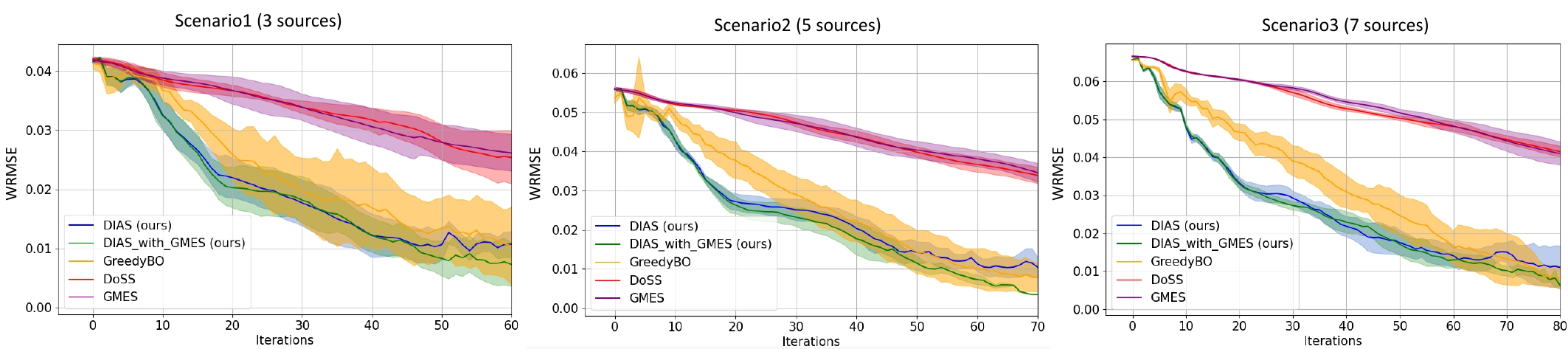}
            \caption{Comparison of the \textit{Weighted Root Mean Squared Error} (WRMSE) trends throughout the different methods across 3 sources (left), 5 sources (middle), and 7 sources (right) scenarios. A faster decrease in the WRMSE value suggests a more rapid convergence of the estimated distribution $\mu(\cdot)$ towards the ground truth distribution $\phi(\cdot)$.}
            \label{fig:rmses}
            \vspace{-2ex}
        \end{figure*}

        We first evaluate \texttt{DIAS} and the baselines by comparing the number of iterations required to find all sources in each of the two scenarios considered. The results are averaged over 30 trials and summarized in Table~\ref{tab:iterations}. It is shown that \texttt{DIAS} and \texttt{DIAS\_with\_GMES} perform the best in each scenario, respectively. The proposed method identifies all sources with speed $3.42$ times faster than that of \texttt{GMES} in Scenario 1, $3.49$ times faster in Scenario 2, and $3.86$ times faster in Scenario 3. As previously mentioned, \texttt{DIAS} identifies potential sources and then alternates between two modes: active sensing (exploration), which involves gathering more information, and source seeking (exploitation), which entails navigating to the potential source. This mechanism significantly enhances our performance compared to the baselines, since all baselines are basically greedy methods, navigating robots to the region with the maximum value of the information function. Thus, the fact that \texttt{DIAS\_with\_GMES} performs well implies that applying our hybrid controller to an existing algorithm yields improved results. Note that \texttt{GreedyBO}, analogous to the proposed framework, partitions the environment into Voronoi cells, whereas \texttt{DoSS} and \texttt{GMES} do not employ this method. This distinction results in \texttt{GreedyBO} outperforming \texttt{DoSS} and \texttt{GMES}, highlighting the importance of constructing and updating Voronoi cells for an efficient search.

        Accurate environment estimation is essential for the robots to locate potential sources. Besides comparing the different approaches in terms of speed, we also look into their environment learning performance. 
        We use \textit{Weighted Root Mean Squared Error} (WRMSE) as another metric, as it gives priority to areas with high $\phi(\cdot)$ values which indicate the presence of sources. 
        Given our interest in those areas,
        WRMSE is defined as:
        \[
        \text { WRMSE }=\sqrt{\frac{\sum_{i=1}^{M}\left[\frac{\left(\mu\left(q\right)-\phi\left(q\right)\right)^2\left(\phi\left(q\right)-\min \phi\left(q\right)\right)}{\max \phi\left(q\right)-\min \phi\left(q\right)}\right]}{M}}, 
        \]
        where $M$ represents the number of data points. Fig. \ref{fig:rmses} illustrates the WRMSE trends throughout the experiment, with the curve representing the mean and the shaded area depicting the variance. \texttt{DIAS} and \texttt{DIAS\_with\_GMES} exhibit the quickest convergence towards the ground truth in both scenarios, indicating superior environment estimation performance. In contrast, \texttt{GreedyBO} trails behind, with \texttt{DoSS} and \texttt{GMES} displaying the slowest convergence, reflecting a less efficient estimation process.
        
         
        \begin{figure*}[tp]
            \centering
            \includegraphics[width=12.5cm]{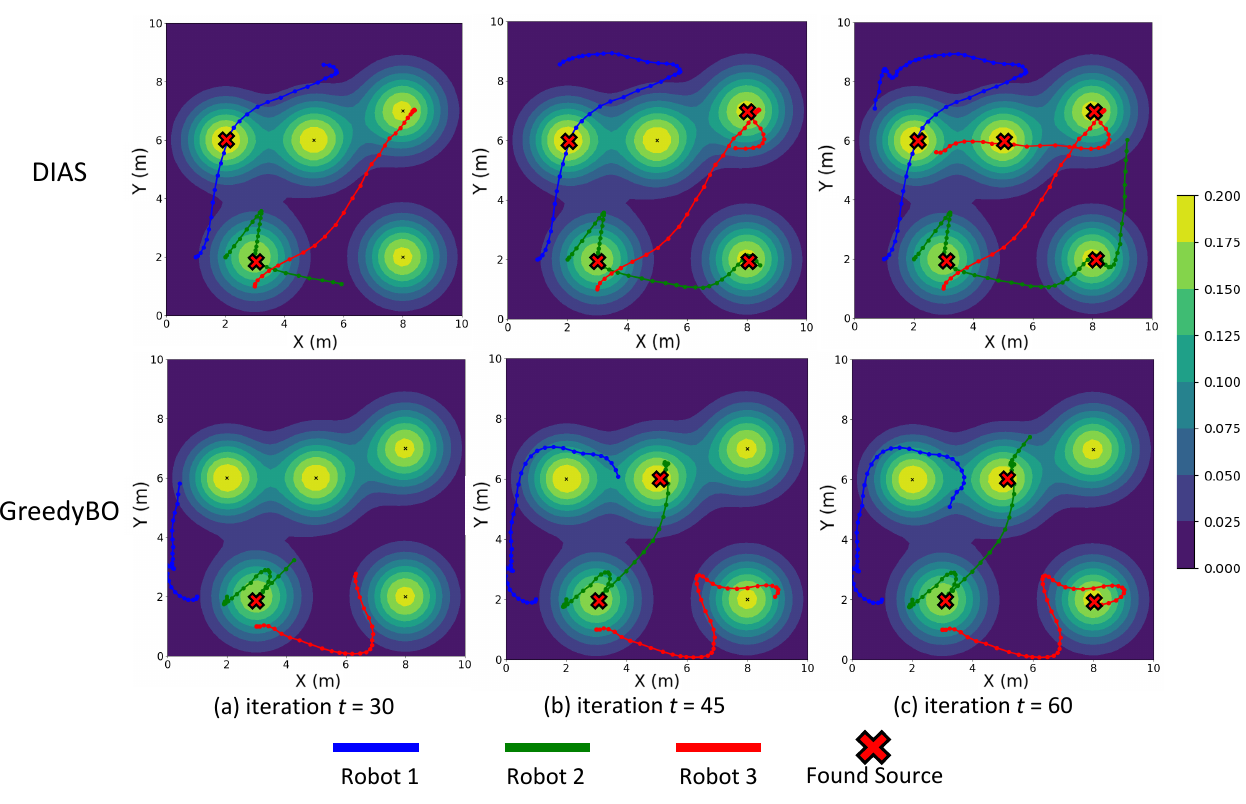}
            \caption{Illustration of three robots searching for five static chemical leakage sources in an unknown area. Each row depicts the trajectory of the robots using \texttt{DIAS} and \texttt{GreedyBO} strategies at $30$, $45$, and $60$ iterations. \texttt{DIAS} successfully identifies all sources within $51$ iterations, whereas \texttt{GreedyBO} requires $106$ iterations. } 
            \label{fig:visual_result}
            \vspace{-2ex}
        \end{figure*}
        
    \subsection{Trajectory Analysis} 

        We now discuss the robot searching in a specific experiment case. Fig. \ref{fig:visual_result} provides the paths of \texttt{DIAS} and \texttt{GreedyBO} in an environment with five sources and captures the progression of searches at iterations \(t = 30\), \(45\), and \(60\). In this case, the robots start from initial positions at \((1, 2)\), \((2, 2)\), and \((3, 1)\), and their search paths are differentiated by color. A source is considered found when a robot comes within \(0.4m\), as established in Sec. \ref{Sec:Problem_setup}. 
        
        In \texttt{DIAS} strategy, the robots exhibit a collaborative search pattern, efficiently covering the space and successfully locating all sources within $51$ iterations. After identifying the sources, the robots continue to search the remaining area, potentially for further exploration or to confirm the absence of additional sources. Conversely, the \texttt{GreedyBO} strategy, which employs a greedy approach by directing each robot to the local maximum of the information function within its Voronoi cell, shows less targeted movement. The second row dedicated to \texttt{GreedyBO} demonstrates that only three sources are located by iteration \(60\), and the overall paths taken by the robots are less direct. The robots' trajectories under the \texttt{GreedyBO} strategy appear to meander more, lacking the directed search pattern observed in the \texttt{DIAS} strategy. This leads to all sources being identified only after \(104\) iterations, indicating a less efficient search process.

    \subsection{Integration with Existing Source Seeking Algorithm}
        \label{SubSec:CombinedMethod}
        A key strength of the proposed framework lies in its ability to enhance existing source seeking algorithms, enabling them to effectively identify all unknown potential sources. As shown above, existing source seeking baselines, such as \texttt{DoSS} and \texttt{GMES} are not well-suited for our source seeking tasks that have an unknown number of sources. This is because they 
        (1) do not use Voronoi tessellation for efficient task allocation, potentially leading to longer and less efficient travel paths, and (2) lack of balance between exploration and exploitation. In contrast, \texttt{DIAS} continuously computes the possibility of the presence of sources using a hybrid controller and efficiently estimates the density function of sources by updating Voronoi cells and employing GP regression. Here, one can integrate existing algorithms into \texttt{DIAS} to take advantage of our framework by replacing the source seeking module with the existing algorithm. As an example, we employ \texttt{GMES} to calculate target points, \( q_{\text{target}} \), instead of using the target computation introduced in Sec. \ref{SubSec:potential_src_identification}. As a result, as shown in Fig. \ref{fig:rmses} and Table \ref{tab:iterations}, \texttt{DIAS\_with\_GMES} significantly outperforms \texttt{GMES}, showing that this integration yields better results.

\section{Conclusion and Future Work}
    In this work, we present \texttt{DIAS}, a novel framework that enables robots to identify all potential sources when the number of sources is unknown, a realistic scenario. The key components of \texttt{DIAS} are twofold. The first is \textit{Environment Estimation}, which involves building and updating Voronoi cells to efficiently divide the environment and approximating the density function of sources based on GP regression. The second is the \textit{Hybrid Controller}, which determines the presence of potential sources by the source identification module and then conducts either exploitation through the source seeking module or exploration via the active sensing module, accordingly. The numerical results demonstrate that the proposed framework surpasses the baselines in both scenarios, whether the number of sources outnumbers or is equal to the number of robots. In addition, the successful integration of the proposed framework with existing source seeking algorithms demonstrates the integrability of \texttt{DIAS}. Future work includes extending our algorithm to handle more practical gas dispersion patterns, such as non-symmetric and turbulent plumes influenced by wind. 

\section*{Acknowledgement}
This work has been supported  by NSF and USDA-NIFA under AI Institute for Resilient Agriculture, Award No. 2021-67021-35329 and Department of Agriculture Award Number  20236702139073.

\printbibliography{}

\end{document}